%% file: root.tex
\documentclass[letterpaper, 10 pt, conference]{ieeeconf}  

\input{settings.tex}

\input{utils.tex}

\IEEEoverridecommandlockouts                              

\title{\LARGE \bf
DistillPose: Lightweight Camera Localization Using Auxiliary Learning
}

\author{Yehya Abouelnaga$^{1}$, Mai Bui$^{2}$ and Slobodan Ilic$^{3}$
\thanks{$^{1}$Yehya Abouelnaga is with Faculty of Informatics,
        Technical University of Munich, 85748 Garching, Germany.
        {\tt\small yehya.abouelnaga@tum.de}. This work was done prior to joining Amazon.}%
\thanks{$^{2}$Mai Bui is with Faculty of Informatics,
        Technical University of Munich, 85748 Garching, Germany.
        {\tt\small mai.bui@tum.de}}%
\thanks{$^{3}$Slobodan Ilic is with Technical University of Munich and Siemens AG, 81739 Munich, Germany.
        {\tt\small slobodan.ilic@siemens.com}}%
}

\begin{document}

\maketitle
\thispagestyle{empty}
\pagestyle{empty}

\begin{abstract}

We propose a lightweight retrieval-based pipeline to predict 6DOF camera poses from RGB images.
Our pipeline uses a convolutional neural network (CNN) to encode a query image as a feature vector.
A nearest neighbor lookup finds the \textit{pose-wise} nearest database image.
A siamese convolutional neural network regresses the relative pose from the nearest neighboring database image to the query image.
The relative pose is then applied to the nearest neighboring absolute pose to obtain the query image's final absolute pose prediction.
Our model is a distilled version of NN-Net \cite{laskar2017camera} that reduces its parameters by 98.87\%, information retrieval feature vector size by 87.5\%, and inference time by 89.18\% without a significant decrease in localization accuracy.

\end{abstract}

\input{sections/introduction.tex}

\input{sections/proposed-approach.tex}
\input{sections/experimental-results.tex}










\bibliographystyle{IEEEtran}
\bibliography{library.bib}

\end{document}

%% file: settings.tex
\usepackage[utf8]{inputenc}
\usepackage[T1]{fontenc}

\usepackage{booktabs} 
\usepackage{caption}
\usepackage[hidelinks]{hyperref} 
\usepackage{ifthen} 
\usepackage{pdftexcmds} 
\usepackage{paralist} 
\usepackage{siunitx} 
\usepackage{multirow} 
\usepackage{array} 
\usepackage{makecell} 
\usepackage{pdfpages} 
\usepackage{adjustbox} 
\usepackage{tablefootnote} 
\usepackage{threeparttable} 
\usepackage{amsfonts}       
\usepackage{nicefrac}       

\usepackage{mathtools}
\usepackage[bottom]{footmisc}
\usepackage{float}
\usepackage{amsmath}
\usepackage{dsfont}
\usepackage{gensymb} 	

\usepackage{hyperref}
\hypersetup{
	colorlinks=true,
	linkcolor=blue,
	filecolor=magenta,      
	urlcolor=cyan,
}

\usepackage{amssymb}
\usepackage{pifont}



%% file: utils.tex
\newcommand{\norm}[1]{\left\lVert#1\right\rVert}

\newcommand{\Loss}{\mathcal{L}}
\newcommand{\Image}{\mathcal{I}}

\newcommand{\Depth}{\mathcal{D}}

\newcommand{\boldK}{\mathbf{K}}

\newcommand{\boldR}{\mathbf{R}}

\newcommand{\boldq}{\mathbf{q}}
\newcommand{\boldt}{\mathbf{t}}

\newcommand{\EuclideanDistance}[2]{\norm{#1 - #2}_2}
\newcommand{\ManhattenDistance}[2]{\norm{#1 - #2}_1}

%% file: sections/introduction.tex

\section{INTRODUCTION}

Camera pose estimation is used in a variety of applications such as autonomous navigation, augmented reality, structure from motion, and simultaneous localization and mapping \cite{shavit2019introduction}.
Given a source of information such as RGB images or depth information, it aims at inferring the 6D camera pose, describing its orientation and position in the scene.

Classically, a feature descriptor (\cite{lowe2004Sift,bay2008Surf,calonder2010BRIEFDescriptor,rublee2011ORB,yi2016LIFT}) would be  used to match 2D points to 3D points.
Assuming the camera is calibrated, solving the Perspective-n-Point (PnP) problem on the correspondences leads to an estimated camera pose \cite{szeliski2010computerVisionBook}.
Such pose estimation approaches are usually very accurate, however, they are often based on prohibitive assumptions: an accurate 3D map of the world is present and a robust key-point descriptor is used.
Further, establishing correct 2D-3D correspondences is a significant challenge as key-point descriptors are sensitive to occlusions, different lighting conditions, and texture.

Several machine learning-based approaches have been proposed that aim to solve this problem by predicting the 3D coordinate for a given 2D point in an image.
\cite{shotton2013SCoReForest} uses random forests \cite{criminisi2013decisionForestsBook} to predict 3D coordinates using RGB images, depth images or both; It does not use the entire image to predict the 3D points but rather takes single pixels as input.
The forest regressors are used in a RANSAC loop \cite{fischler1981randomRansac} where a pose is refined iteratively.
\cite{brachmann2017Dsac} develops a differentiable RANSAC \cite{fischler1981randomRansac} pipeline for pose estimation.
Such methods yield very impressive results.
Their flaw is the need for scene-specific training--- thus, requiring a new predictor to be trained to adopt to any new scene.

\input{figures/triplets-sample.tex}

With the recent advancements in deep learning, the pose estimation task has been formulated as a regression problem where a 7-dimensional vector representing the absolute pose would be regressed.
\cite{kendall2015posenet} introduced the first absolute pose regression model on top of CNNs.
They use a GoogLeNet \cite{szegedy2015goingGoogLeNet} backbone to transform scene images to feature vectors;
A fully-connected layer regresses both absolute translations and rotation unit quaternions.
\cite{kendall2016modellingBayesianPoseNet,kendall2017geometricLossFunctions,walch2017LSTMPose} proposed small variations to the PoseNet \cite{kendall2015posenet} architecture to reduce median localization error.
Pose regression from RGB images is much faster than 2D-3D correspondence-based approaches.
However, it suffers from two drawbacks: 1) The network is scene-specific and requires a significant amount of offline training and 2) it lacks in performance compared to correspondence-based approaches.
NN-Net \cite{laskar2017camera} address these limitations by decomposing the \textit{absolute pose regression} problem into two components: information retrieval and relative pose regression.
The \textit{information retrieval} component finds a reference image (with corresponding absolute pose) that is closest to the query image (See Fig. \ref{figure:quadruplets-batch}).
The \textit{relative pose regression} component then regresses the relative pose and applies it to the reference image's absolute pose.
RelocNet \cite{balntas2018relocnet} correlates learned features to changes in the viewing camera frustum to improve nearest neighbor retrieval.
CamNet \cite{ding2019camnet} introduces auxiliary learning in a coarse-to-fine framework to significantly reduce median localiztion error.
This setup is able to generalize to new and unseen scenes-- as relative pose regression is not dependent on the scene's world coordinates.
It also outperforms most absolute pose regression solutions.
These improvements come at a computational and storage cost; Running deep siamese networks on edge devices (i.e. mobile phones, robots) is slow.
Using five nearest neighbors (as in \cite{ding2019camnet} and \cite{laskar2017camera}) requires five inference steps.
Storing thousands of database images requires plenty of on-device memory.
This renders on-device camera localization in realtime unpractical for many robotic applications.
In this paper, our contributions are twofold: we propose a novel layerwise training approach that enforces earlier layers to learn better pose-aware embeddings, and we use angle-based and frustum-based auxiliary losses in a fine-tuning step to:

\begin{itemize}
    \item Use a single nearest neighbor (instead of 5 in NN-Net \cite{laskar2017camera}) and remove the need for a RANSAC pose filtering algorithm,
    \item Reduce the model size by 98.87\% (from 22.3M parameters in NN-Net \cite{laskar2017camera} to 248K parameters),
    \item Reduce the information retrieval feature vector size by 87.5\% (from a 512-dimensional vector to a 64-dimensional one),
    \item Reduce our inference time by 89.18\% (from 37ms to 4ms),
    \item Outperform NN-Net \cite{laskar2017camera} on information retrieval while retaining the performance on pose regression.
\end{itemize}

\input{figures/inference-pipeline.tex}

%% file: figures/triplets-sample.tex
\begin{figure}
  \centering
  \includegraphics[width=0.48\textwidth]{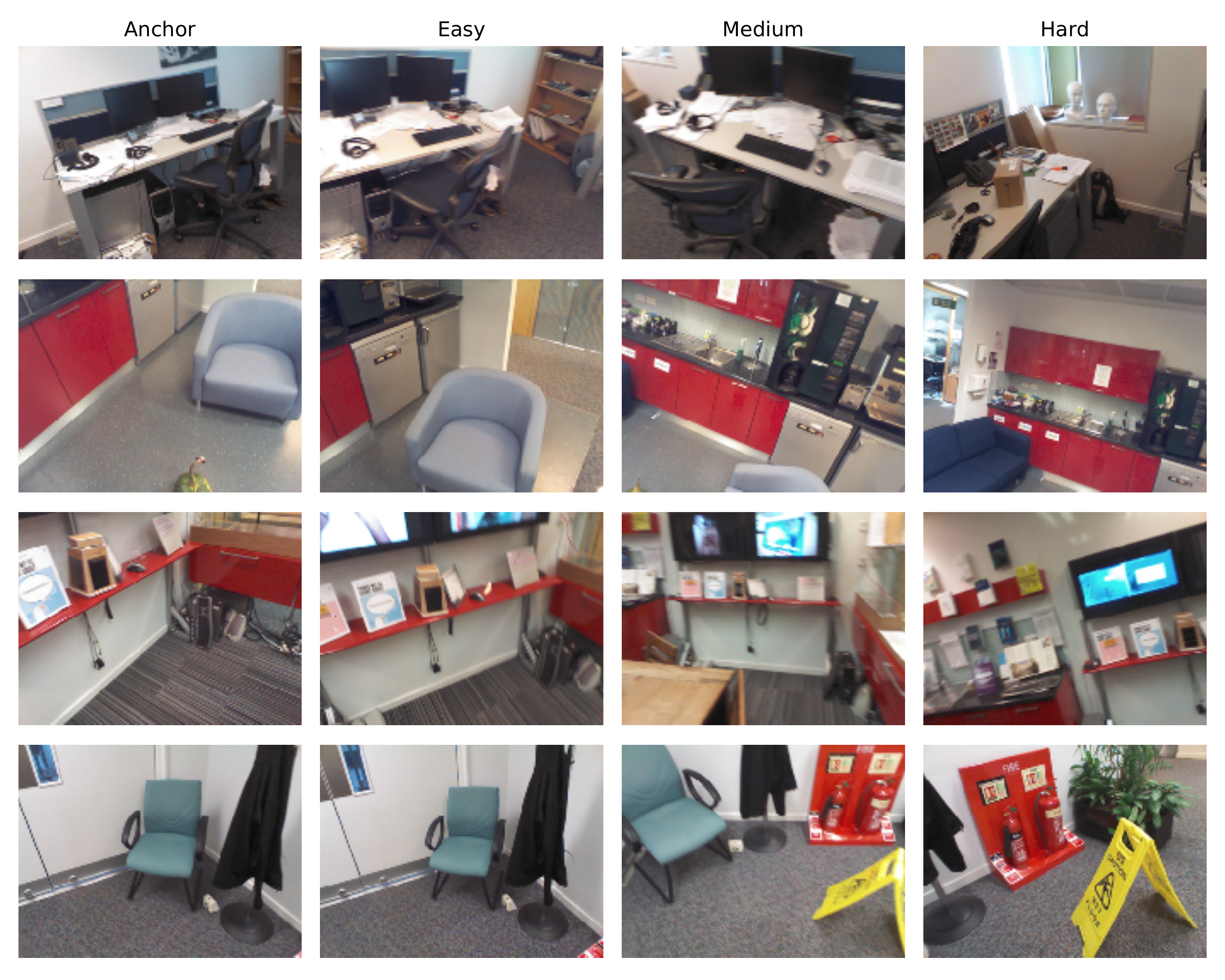}
  \caption{
    \textit{Relative Pose Regression.} Our relative pose regression network learns to predict the change in camera pose from a reference image to a query image. We train our regressor on pairs with easy, medium and hard changes in pose \cite{ding2019camnet}.
    }
    \label{figure:quadruplets-batch}
\end{figure}

%% file: figures/inference-pipeline.tex

\begin{figure}
    \centering
    \includegraphics[width=0.48\textwidth]{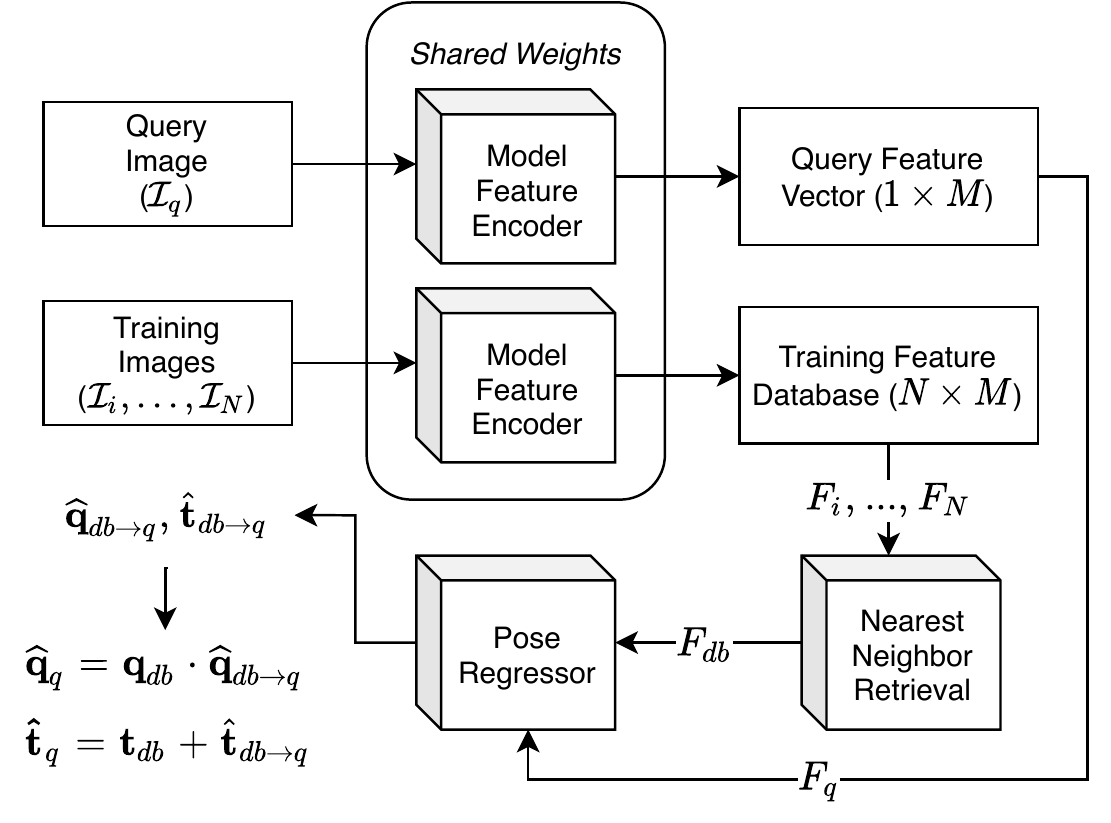}
    \caption{Our End-to-End Camera Localization Pipeline.}
  \label{figure:test-pipeline}
\end{figure}

%% file: sections/proposed-approach.tex
\input{figures/layerwise-model.tex}

\section{Proposed Approach}
Throughout the literature, multi-modal learning has shown to persistently aid in pose estimation tasks \cite{ding2019camnet,balntas2018relocnet,saha2018AnchorNet};
We leverage this assumption and propose DistillPose. Our model consists of two stages, 1) layerwise training that incorporates pose information into a lightweight neural network for pose regression and 2) fine-tuning the model for improved pose-aware feature learning. 
\subsection{Layerwise Training}
\label{section:layerwise-training}
We begin with the assumption that enforcing a pose loss on each residual layer of the ResNet34 would guide the network to learn pose-aware information.
We train all layers simultaneously (i.e. the full model in Fig. \ref{figure:layerwise-model}).
After the first convolutional block, ResNet34 \cite{he2016resnet} has four residual blocks;
We apply a global average pooling on outputs of each residual block-- resulting in $64$-, $128$-, $256-$ and $512$-dimensional vectors, dubbed here as $L_1$, $L_2$, $L_3$ and $L_4$, respectively.
For each residual block, we learn an independent relative pose regression function: $P_1$, $P_2$, $P_3$, and $P_4$.
Each relative pose regression head learns to predict the relative pose from four different feature layers in ResNet34.
Pose regression heads are fed the concatenated feature vectors from two images (a database and a query image);
Thus, $P_1$, $P_2$, $P_3$, and $P_4$ expect $128$-, $256$-, $512$- and $1024$-dimensional input vectors to regress a relative pose, trained on the following loss function

\[
  \Loss_{PL} =
  \sum_{i = 1}^{4}
  \ManhattenDistance{
    \hat{\boldt}^{(i)}_{db \rightarrow q}
  }{
    \boldt^{(i)}_{db \rightarrow q}
  } 
  + \beta
  \ManhattenDistance{
    \hat{\boldq}^{(i)}_{db \rightarrow q}
  }{
    \boldq^{(i)}_{db \rightarrow q}
  },
\]
where $\boldt_{db \rightarrow q} = \boldt_{q} - \boldt_{db}$ and $\boldq_{db \rightarrow q} = \boldq^{-1}_{db} \boldq_{q}$ are the ground truth relative translation and rotation, represented as a quaternion, and $\hat{\boldt}_{db \rightarrow q}$ as well as $\hat{\boldq}_{db \rightarrow q}$ the predictions.
Regressed relative poses are applied to the database image's absolute pose (as in Fig. \ref{figure:test-pipeline}) to obtain the query image's absolute translation \( \mathbf{\hat{t}}_{q} = \mathbf{t}_{db} + \hat{\mathbf{t}}_{db \rightarrow q} \) and rotation \(\mathbf{\hat{q}}_{q} = \mathbf{q}_{db} \cdot \hat{\mathbf{q}}_{db \rightarrow q}\).

\subsection{Learning frustum-based embeddings}
\label{section:learning-frustum-based-embeddings}
CamNet \cite{ding2019camnet} and RelocNet \cite{balntas2018relocnet} use a frustum loss to enforce a minimum frustum overlap between images with similar feature vectors.
In a similar fashion, we fine-tune our distilled version of the model (i.e. short version in Fig. \ref{figure:layerwise-model}) to improve the pipeline's information retrieval and relative pose regression.
We assume that we have a pair of images $\Image_a$ and $\Image_b$ with known absolute rotations ($\boldR_a$ and $\boldR_b$), absolute translations ($\boldt_a$ and $\boldt_b$) and depth maps ($\Depth_a$ and $\Depth_b$).
Given that the camera is calibrated and the intrinsics matrix $\boldK$ is known a priori, we can project a pixel $X^{(a)}_{j}$ from $\Image_a$ to $\Image_b$ to obtain $X^{(b)}_{j}$ as follows:

\[
  X^{(b)}_{j} = \boldK \cdot (\boldR_b^T \boldR_a \boldK^{-1} X^{(a)}_{j} \Depth_a + \boldt_a - \boldt_b)
\]

$X^{(b)}_{j}$ is the pixel projected in the second image.
The frustum overlap $\theta_1$ is measured as the percentage of pixels from the first image $\Image_a$ that are re-projected inside the area of the second image $\Image_b$.
The frustum distance $d_1 = 1 - \theta_1$ can be used to measure the distance between two camera frustums.
$d_2$ describes the frustum distance for pixels re-projected from $\Image_b$ to $\Image_a$.

\subsubsection{Predicting frustum distances}
\label{section:predicting-frustum-distances}

A possible way to learn a frustum distance-aware embedding space is to use a fully connected layer to predict frustum distance as in \cite{ding2019camnet}.
A fully connected layer learns to predict the bilateral frustum distances $\hat{d}_1$ and $\hat{d}_2$.
An $L_1$-based loss could then be used to enforce the distance the function:

\[
  \Loss_{PF} = \ManhattenDistance{\hat{d}_1}{d_1} + \ManhattenDistance{\hat{d}_2}{d_2}
\]

\subsubsection{Enforcing frustum distances directly on embeddings}
\label{section:enforcing-frustum-distance}

Instead of predicting bilateral frustum distances (as in \cite{ding2019camnet}), we could enforce the frustum distance to be the distance between two image feature vectors directly (as in \cite{balntas2018relocnet}).
Assume that $\phi(a)$ and \(\phi(b)\) are the feature vectors for a pair of images \(\Image_a\) and \(\Image_b\);
The following loss function enforces the embedding space to reflect the pair frustum overlap.

\[
  \Loss_{EF} = \ManhattenDistance{d_1}{
    \EuclideanDistance{\phi(a)}{\phi(b)}
  } + \ManhattenDistance{d_2}{
    \EuclideanDistance{\phi(a)}{\phi(b)}
  }
\]

\subsubsection{Enforcing frustum distances using triplet loss}
\label{section:frustum-triplet-loss}

While CamNet \cite{ding2019camnet} learned to predict the frustum distances between pairs of images, they also constructed a set of easy, medium and hard pairs for triplet loss-based training.
All pairs share the same anchor image $\mathcal{I}_a$.
The hard pairs have bilateral frustum overlaps less than 25\% and greater than 5\%.
Alternatively, instead of learning frustum distances, we could teach the network to pull pairs of images with high frustum overlap together while pushing pairs with lower frustum overlap farther apart.
To construct a triplet, we could use easy and hard pairs--- which could be combined in the following triplet $(\Image_a, \Image_e, \Image_h)$ (where $\Image_a$, $\Image_e$, $\Image_h$ refer to the anchor, positive and negative images, respectively).

\[
  \mathcal{L}_{FTL} = \left[
    m + 
    \EuclideanDistance{\phi(a)}{\phi(e)}
    -
    \EuclideanDistance{\phi(a)}{\phi(h)}
  \right]_{+}
\]

\(m\) is the triplet loss margin; \(\left[\cdot\right]_+ = \max(\cdot, 0)\) is the Hinge loss.
We could also use two triplet losses: one combines easy and hard pairs and the other combines medium and hard pairs.

\subsection{Learning angle-based embeddings}
\label{section:learning-angle-based-embeddings}

For a pair of images $\Image_a$ and $\Image_b$ with unit quaternions \(\boldq_a\) and \(\boldq_b\), the angular distance \(\alpha\) between the unit quaternions could be computed as follows

\[
  \alpha = 2 \arccos \left( \lvert \boldq_a^T \boldq_b \rvert \right) / \pi.
\]

CamNet \cite{ding2019camnet} showed that imposing an angular distance constraint helps improve the image retrieval and pose regression performance.
We experiment with imposing an angular distance-based constraint and analyze its impact on the information retrieval and pose regression components.
We are interested in comparing different approaches of imposing the angular distance constraint.

\subsubsection{Predicting angular distance}
\label{section:predicting-angular-distance}

We use a fully connected layer that predicts the angle ($\hat{\alpha}$) between two images $\Image_a$ and $\Image_b$.
We encode our images into the corresponding feature vectors $\phi(a)$ and \(\phi(b)\).
An $L_1$-based loss is then used to learn the angular distance

\[
  \Loss_{PA} = \ManhattenDistance{\hat{\alpha}}{\alpha}.
\]

\subsubsection{Enforcing angular distances directly on embeddings}
\label{section:enforcing-angular-distance}

Instead of predicting angular distances (as in \cite{ding2019camnet}), we could enforce the angular distance to be the distance between two image feature vectors directly.
Assume that $\phi(a)$ and \(\phi(b)\) are the feature vectors for a pair of images \(\Image_a\) and \(\Image_b\);
The following loss function enforces the embedding space to reflect the pair angular distance

\[
  \Loss_{EA} = \ManhattenDistance{\alpha}{
    \EuclideanDistance{\phi(a)}{\phi(b)}.
  }
\]

\subsubsection{Enforcing angular distances using triplet loss}
\label{section:angular-triplet-loss}

In the frustum distance loss, we showed that the CamNet-generated \cite{ding2019camnet} quadruplets of anchor, easy, medium and hard images could be used to learn a frustum loss by using easy and hard pairs (or, medium and hard pairs).
We also realize that the easy and medium pairs have non-overlapping angular distances (See Fig. \ref{figure:quadruplets-distribution});
The medium pairs have angular distance greater than 60\%--- while the easy pairs have an angular distance less than 30\%.
In other words, we could teach the network to pull together pairs of images with low angular distances (i.e. the easy pairs) while pushing pairs with higher angular distance (i.e. medium pairs) farther apart.
To construct a triplet, we use easy and medium pairs--- which could be combined in the following triplet $(\Image_a, \Image_e, \Image_m)$ (where $\Image_a$, $\Image_e$, $\Image_m$ refer to the anchor, positive and negative images, respectively) and train on the following loss function

\[
  \mathcal{L}_{ATL} = \left[
    m + 
    \EuclideanDistance{\phi(a)}{\phi(e)}
    -
    \EuclideanDistance{\phi(a)}{\phi(m)}
  \right]_{+}.
\]




\subsection{Multi-Instance Learning}
\label{section:multi-instance-learning}
When using angle-based and frustum-based auxiliary losses in a fine-tuning step, the network is learning two different functions simultaneously.
Combining different functions is tricky as different loss functions do have different magnitudes and contribute to the model's learning outcome disproportionately.
\cite{kendall2015posenet} dealt with multi-instance learning weights by manually assigning weights to different loss terms.
In our fine-tuning step, we start with combining different loss functions while assuming equal loss term weighting as follows:

\[
  \Loss_{PL+PF} = \Loss_{PL} + \Loss_{PF}
\]

However, we observe that, in a fine-tuning step, fine-tuned loss functions (i.e. \(\Loss_{PF}, \Loss_{EF}, \Loss_{EA}\)) would be initially 6-10 times the magnitude of the pose loss (\(\Loss_{PL}\)).
\cite{kendall2017uncertaintiesHomoscedastic} introduced \textit{homoscedastic loss}, a loss term-weighting technique, that has been successfully applied in \cite{kendall2017geometricLossFunctions}.
We use the homoscedastic loss term weighting as follows:

\[
	\Loss_{PL+PF+H} = \frac{\Loss_{PL}}{exp(\hat{\beta})} + \hat{\beta} + \frac{\Loss_{PF}}{exp(\hat{\gamma})} + \hat{\gamma}
\]

\(\hat{\gamma}\) and \(\hat{\beta}\) are both learnable hyper-parameters that are robust to initialization (as per \cite{kendall2017uncertaintiesHomoscedastic}).



%% file: figures/layerwise-model.tex
\begin{figure*}[h]
  \centering
  \includegraphics[width=\textwidth]{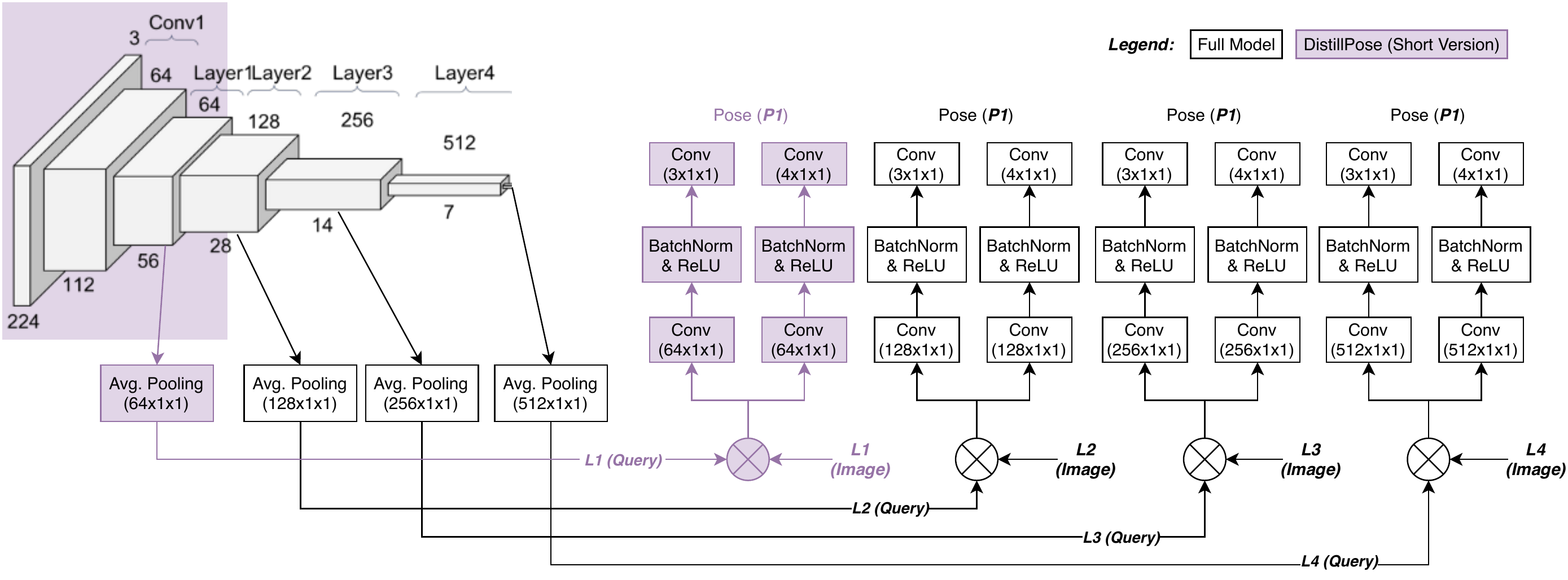}
  \caption{Overview of DistillPose, that leverages layerwise relative pose regression to produce a lightweight pose-aware model.}
  \label{figure:layerwise-model}
\end{figure*}

%% file: sections/experimental-results.tex

\input{figures/triplets-stats.tex}
\input{figures/laskar-stats.tex}

\section{Experimental Results}

\subsection{Dataset and Image Pair Construction}
We use the Seven Scenes dataset \cite{glocker2013SevenScenes} to train and test our different model variations.
In test time, we use all testing frames as test query images and predict a final absolute pose for each.
For training a relative pose estimation model, we require a dataset of image pairs.
We experimentally found this task to be challenging, as the the choice of pairing significantly impacts the inference performance.
If a dataset of consecutive pairs is used, the relative pose estimation model learns to only predict pairs with small changes in pose.
Another problem with using consecutive images in a sequence as training pairs is model collapse;
Oftentimes, the change in pose in consecutive pairs is very small.
This could lead the network to learn "no change in pose" as the optimal solution (i.e. always predicting zeros or very small values irrespective of the given image pairs).

\textbf{Image Pair Construction.}
Much work has already been invested in constructing reasonable set of image pairs for relative pose estimation;
We use the pairs generated by both CamNet \cite{ding2019camnet} and NN-Net \cite{laskar2017camera}.

\subsubsection{CamNet Quadruplets \cite{ding2019camnet}}
\label{section:camnet-quadruplets}
\cite{ding2019camnet} constructed a dataset of three pairs of (anchor, easy), (anchor, medium) and (anchor, hard) images.
Easy pairs have $>40\%$ frustum overlap with angular distance $<30\%$.
Medium pairs have $>30\%$ frustum overlap with angular distance $>60\%$.
Hard pairs have a frustum overlap between 5\% and 25\% and no rotational constraints.
Fig. \ref{figure:quadruplets-distribution} shows each pair's translational, rotational and frustum distances.
Fig. \ref{figure:quadruplets-batch} shows a sample of Seven Scenes quadruplets.

\subsubsection{NN-Net Pairs \cite{laskar2017camera}}
\label{section:nn-net-pairs}
\cite{laskar2017camera} followed a simpler strategy;
For each training frame in the training data, you would find a corresponding image that would have ``sufficiently overlapping field of view''.
\cite{laskar2017camera} did not mention what counts as sufficiently overlapping field of view.
Fig. \ref{figure:laskar-pairs-distribution} shows the NN-Net \cite{laskar2017camera} training pairs translational, rotational and frustum distances.

\input{tables/ablation-study.tex}

\subsection{Implementation Details}

\textbf{Pre-training.}
ResNet34's weights are initialized from a pre-trained model on ImageNet classification task \cite{deng2009imagenet}.
We run the layerwise training step proposed in Sec. \ref{section:layerwise-training}.
We use NN-Net training pairs (Sec. \ref{section:nn-net-pairs}).
For each pair, we obtain feature vectors from the four residual blocks (i.e. \(L_1, L_2, L_3\) and \(L_4\)) for both database and query image.
We concatenate database and query image feature vectors per residual block and regress the layerwise pose using residual block-specific regression heads (i.e. \(P_1, P_2, P_3\) and \(P_4\)).
Our pose regression heads are randomly initialized.
We train for 300 epochs using an Adam optimizer with an initial 0.0001 learning rate and a 128 batch size.

\textbf{Fine-tuning.}
In this step, we remove all residual blocks and pose regressors except the first residual block (i.e. \(L_1\)) and its pose regressor (i.e. \(P_1\)).
We initialize the distilled model with weights from the previous pre-training step.
We use auxiliary losses (Sec. \ref{section:learning-frustum-based-embeddings} and Sec. \ref{section:learning-angle-based-embeddings}) alongside the pose loss (Sec. \ref{section:layerwise-training}).
We use the CamNet quadruplets (Sec. \ref{section:camnet-quadruplets}) for training.
For pose loss, across all variants, we use easy pairs.
For predicting and enforcing frustum distances (i.e. ``EF'' and ``PF'' variants; See Table \ref{table:information-retrieval-results}), we use easy, medium and hard pairs.
For predicting and enforcing angular distances (i.e. ``EA'' and ``PA'' variants), we use easy and medium pairs.
For angular-based triplet loss (i.e. ``ATL''), we use easy and medium pairs.
For frustum-based triplet loss (i.e. ``FTL''), we use easy and hard pairs.
We train for 75 epochs using an SGD optimizer with an initial 0.0001 learning rate and a 128 batch size.

\textbf{Training Environment.}
We resize training images to \(224 \times 224\).
Similar to \cite{laskar2017camera,ding2019camnet}, we perform flip, random gaussian blur and color jittering with a 0.4 threshold.
We trained all our networks on a 4-vCPU {\tt g4dn.xlarge} AWS EC2 spot instance with a single NVIDIA T4 GPU.
Our model inference time has been measured on this spot instance.
We measured RANSAC pose filtering (in NN-Net \cite{laskar2017camera}) compute time in a 4-vCPU virtual box on MacOS.

\subsection{Evaluation}

In Table \ref{table:information-retrieval-results}, we measure the translational and rotational distance between test images and their corresponding nearest neighbor.
This shows how close the selected nearest neighbor is.
A good embedding space would pull images of similar poses closer together (thus reduce the median translation and rotation error).
In Table \ref{table:pose-regression-results}, we measure the translational and rotational distance between test images and their predicted absolute poses.
Initially, we notice that earlier layers (i.e. \(L_1\) and \(L_2\)) in our pre-trained model outperform latent layers on both the information retrieval and the pose regression tasks.
In Table \ref{table:pose-regression-results}, all model variants with auxiliary losses (except ``L1 (PL+EA+H)'') outperform the original layerwise pre-trained model (i.e. ``L1 (PL)'') on rotation regression.
We notice that, on the ``PL+EF'', ``PL+PA'' and ``PL+PF'' model variants, using a homoscedastic loss reduces both translational and rotational errors.
Finally, triplet loss-based variants outperform all other variants without using a homoscedastic loss.
Our best performing model variant ``PL+PA+H'' (followed closely by the angle-based triplet loss variant, ``PL+ATL'') learns to regress the angular distance and combines angular distance loss term with pose loss term using a homoscedastic loss. 

When comparing to NN-Net \cite{laskar2017camera}, a few things stand out;
On the information retrieval task, our earlier layers are already on parity with NN-Net \cite{laskar2017camera}.
However, on the pose regression task, NN-Net \cite{laskar2017camera} outperforms all our pre-trained models, especially on rotation regression.
When using auxiliary learning, our angle-based model variants (i.e. ``PL+ATL'' and ``PL+PA+H'') outperform NN-Net \cite{laskar2017camera} on information retrieval while retaining a similar performance on pose regression.
Our distilled NN-Net model (DistillPose) uses a single nearest neighbor (instead of 5 in NN-Net \cite{laskar2017camera}).
Without a significant increase in median localization errors (i.e. ``PL+PA+H'' varient with $0.23m\,9.85\degree$ vs. NN-Net's $0.21m\,9.30\degree$), our solution reduces the model size by 98.87\% (from 22.3M parameters in NN-Net to 248K parameters), information retrieval feature vector size by 87.5\% (from a 512-dimensional vector to a 64-dimensional one), and inference time by 89.18\% (from 37ms to 4ms).

\section{Conclusion}

In this paper, we propose a novel layerwise training approach that enforces earlier layers to learn better pose-aware embeddings.
In addition, we examine different angle-based and frustum-based auxiliary losses in a fine-tuning step to learn a better pose regression function using a smaller version of ResNet34.
Our distilled NN-Net model (DistillPose) uses a single nearest neighbor (instead of 5 in NN-Net \cite{laskar2017camera})--- thus, removing the need for a RANSAC pose filtering algorithm.
Without a significant increase in median localization errors, our solution reduces the model size, information retrieval feature vector size and inference time in comparison to the state of the art.

%% file: figures/triplets-stats.tex
\begin{figure}
  \centering
  \includegraphics[width=0.48\textwidth]{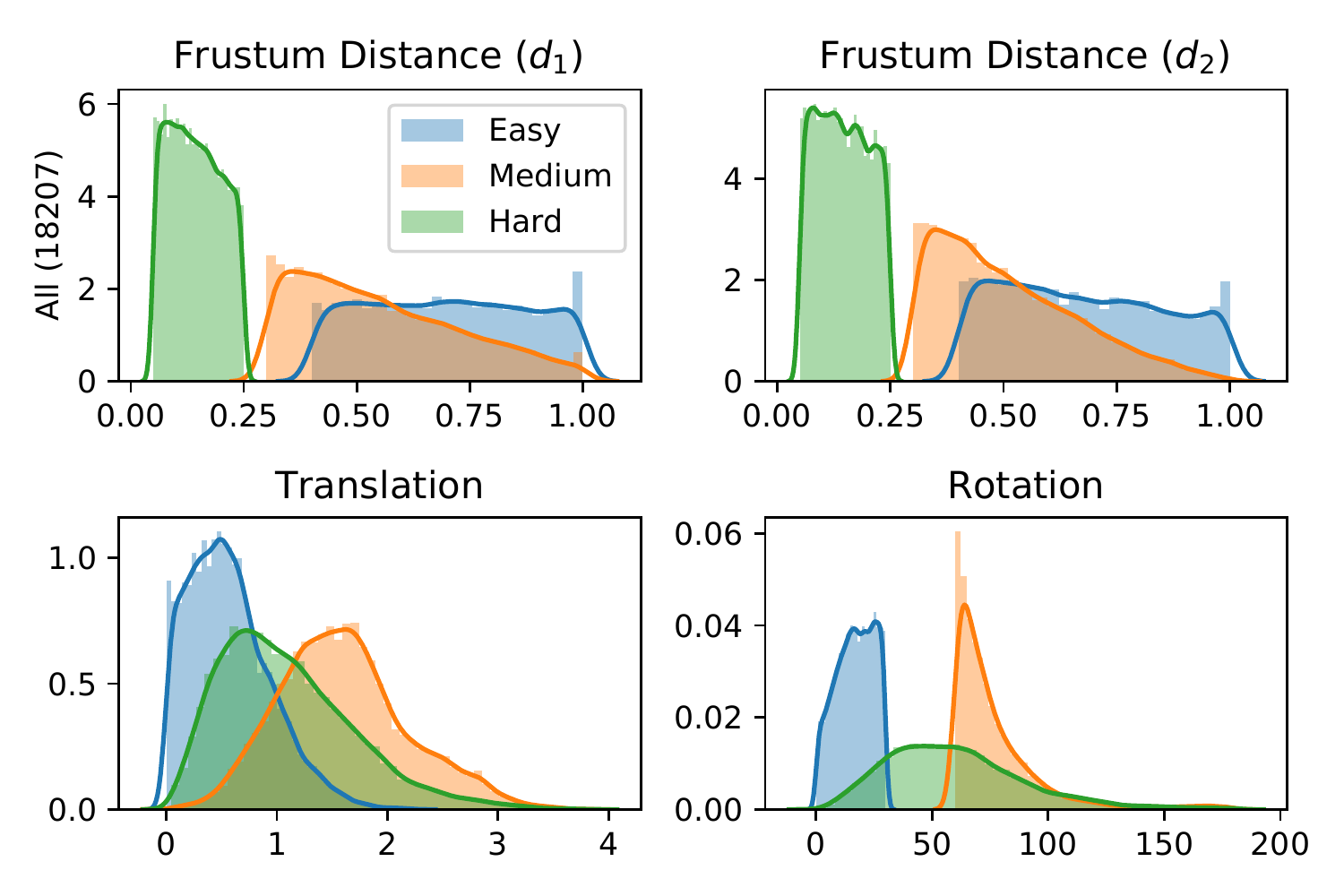}
  \caption{Quadruplets Distribution.}
  \label{figure:quadruplets-distribution}
\end{figure}

%% file: figures/laskar-stats.tex
\begin{figure}
  \centering
  \includegraphics[width=0.48\textwidth]{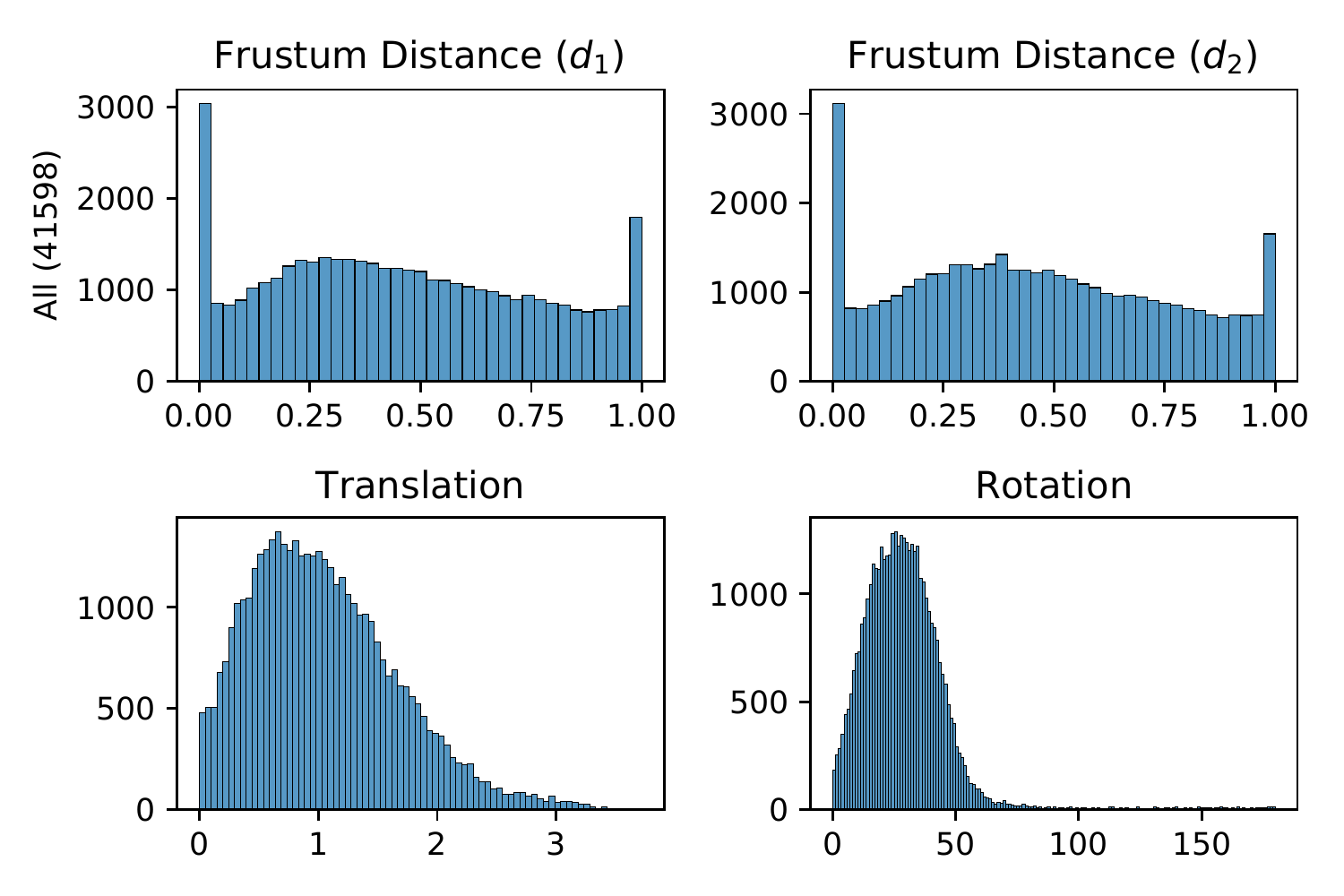}
  \caption{NN-Net Pairs Distribution.}
  \label{figure:laskar-pairs-distribution}
\end{figure}

%% file: tables/ablation-study.tex
\begin{table*}
  \centering
  \caption{
    \textbf{Information Retrieval Median Localization Error.}
    ``PL'' refers to the L1 pose loss (Sec. \ref{section:layerwise-training}).
    ``EA'' refers to enforcing the angular distance directly on the embedding space (Sec. \ref{section:enforcing-angular-distance}).
    ``EF'' refers to enforcing the bilateral frustum distance directly on the embedding space (Sec. \ref{section:enforcing-frustum-distance}).
    ``PA'' refers to predicting the angular distance (Sec. \ref{section:predicting-angular-distance}).
    ``PF'' refers to predicting the bilateral frustum distances (Sec. \ref{section:predicting-frustum-distances}).
    ``H'' refer to model variants where a homoscedastic loss has been used (Sec. \ref{section:multi-instance-learning}).
  }
  \label{table:information-retrieval-results}
  \resizebox{\textwidth}{!}{
    \begin{tabular}{l|ccccccc|c}
      \toprule
      Scene &                   Chess &                  Office &                           Stairs &                   Heads &                    Fire &                 Pumpkin &              Redkitchen &                 Average \\
      \midrule
      NN-Net \cite{laskar2017camera} &           0.25m, 12.97\degree &           0.35m, 12.21\degree &  \textbf{0.33}m, \textbf{13.89}\degree &           0.25m, 18.65\degree &  \textbf{0.34}m, 15.62\degree &           0.43m, 15.47\degree &           0.38m, 15.00\degree &           0.33m, 14.83\degree \\
      L4 (PL)                        &           0.26m, 13.90\degree &           0.34m, 15.83\degree &                    0.52m, 16.83\degree &           0.45m, 20.33\degree &           0.52m, 16.82\degree &           0.40m, 17.93\degree &           0.41m, 16.53\degree &           0.41m, 16.88\degree \\
      L3 (PL)                        &           0.26m, 13.78\degree &           0.33m, 16.00\degree &                    0.46m, 18.86\degree &           0.44m, 20.21\degree &           0.46m, 17.45\degree &           0.38m, 16.84\degree &           0.39m, 16.15\degree &           0.39m, 17.04\degree \\
      L2 (PL)                        &  \textbf{0.23}m, 12.19\degree &  \textbf{0.30}m, 13.79\degree &                    0.36m, 15.66\degree &           0.27m, 17.76\degree &           0.36m, 16.06\degree &  \textbf{0.31}m, 14.04\degree &  \textbf{0.33}m, 14.45\degree &  \textbf{0.31}m, 14.85\degree \\
      L1 (PL)                        &           0.27m, 12.79\degree &           0.36m, 12.45\degree &                    0.45m, 15.76\degree &           0.23m, 18.25\degree &           0.41m, 14.13\degree &           0.38m, 11.36\degree &           0.37m, 13.29\degree &           0.35m, 14.00\degree \\
      L1 (PL+EA)                     &           0.28m, 11.26\degree &           0.39m, 11.76\degree &                    0.58m, 14.53\degree &           0.27m, 15.81\degree &           0.46m, 14.98\degree &           0.42m, 11.25\degree &           0.41m, 11.91\degree &           0.40m, 13.07\degree \\
      L1 (PL+EA+H)                   &  0.31m, \textbf{10.74}\degree &           0.43m, 11.99\degree &           0.58m, \textbf{13.89}\degree &  0.29m, \textbf{13.83}\degree &           0.53m, 17.46\degree &           0.43m, 11.31\degree &  0.44m, \textbf{11.67}\degree &           0.43m, 12.99\degree \\
      L1 (PL+EF)                     &           0.31m, 14.50\degree &           0.41m, 14.71\degree &                    0.69m, 14.81\degree &           0.34m, 19.26\degree &           0.45m, 14.99\degree &           0.44m, 13.61\degree &           0.44m, 15.50\degree &           0.44m, 15.34\degree \\
      L1 (PL+EF+H)                   &           0.29m, 13.99\degree &           0.38m, 14.17\degree &                    0.57m, 14.09\degree &           0.29m, 17.34\degree &           0.44m, 15.34\degree &           0.40m, 12.23\degree &           0.40m, 14.97\degree &           0.40m, 14.59\degree \\
      L1 (PL+PA)                     &           0.26m, 12.19\degree &           0.34m, 11.98\degree &                    0.48m, 15.50\degree &           0.23m, 15.51\degree &           0.40m, 13.38\degree &           0.40m, 11.10\degree &           0.35m, 12.97\degree &           0.35m, 13.23\degree \\
      L1 (PL+PA+H)                   &           0.25m, 12.36\degree &           0.35m, 12.40\degree &                    0.40m, 14.95\degree &  \textbf{0.21}m, 16.41\degree &           0.41m, 13.66\degree &           0.37m, 12.16\degree &           0.34m, 12.70\degree &           0.33m, 13.52\degree \\
      L1 (PL+PF)                     &           0.26m, 13.00\degree &           0.35m, 12.19\degree &                    0.45m, 15.94\degree &           0.25m, 17.21\degree &  0.44m, \textbf{13.33}\degree &           0.39m, 11.47\degree &           0.35m, 13.29\degree &           0.36m, 13.78\degree \\
      L1 (PL+PF+H)                   &           0.27m, 12.29\degree &           0.34m, 12.37\degree &                    0.36m, 16.38\degree &  \textbf{0.21}m, 18.21\degree &           0.41m, 14.30\degree &           0.39m, 11.55\degree &           0.34m, 13.05\degree &           0.33m, 14.02\degree \\
      L1 (PL+ATL)                    &           0.27m, 11.66\degree &  0.36m, \textbf{11.64}\degree &                    0.55m, 14.29\degree &           0.23m, 15.99\degree &           0.44m, 13.77\degree &  0.38m, \textbf{10.82}\degree &           0.40m, 12.15\degree &  0.38m, \textbf{12.90}\degree \\
      L1 (PL+FTL)                    &           0.30m, 11.34\degree &           0.37m, 12.11\degree &                    0.47m, 15.40\degree &           0.29m, 16.84\degree &           0.44m, 14.54\degree &           0.43m, 12.36\degree &           0.41m, 12.91\degree &           0.39m, 13.64\degree \\
      \bottomrule
    \end{tabular}
  }
  \bigskip
  \caption{
    \textbf{Pose Regression Median Localization Error.}
    L1-prefixed modes (ours) are only using the feature vectors of the first residual block in ResNet34 \cite{he2016resnet}.
    We use \(k=1\) nearest neighbor (vs. 5 neighbors in \cite{laskar2017camera}) and no RANSAC pose filtering resulting in an 89.18\% reduction in inference time (from 37ms to 4ms).
  }
  \label{table:pose-regression-results}
  \resizebox{\textwidth}{!}{
    \begin{tabular}{l|ccccccc|c}
      \toprule
      Scene &                           Chess &                          Office &                           Stairs &                   Heads &                             Fire &                Pumpkin &                      Redkitchen &                         Average \\
      \midrule
      NN-Net \cite{laskar2017camera} &  \textbf{0.13}m, \textbf{6.46}\degree &  \textbf{0.21}m, \textbf{7.35}\degree &  \textbf{0.27}m, \textbf{11.82}\degree &  \textbf{0.14}m, 12.34\degree &  \textbf{0.26}m, \textbf{12.72}\degree &  0.24m, \textbf{6.35}\degree &  \textbf{0.24}m, \textbf{8.03}\degree &  \textbf{0.21}m, \textbf{9.30}\degree \\
      L4 (PL)                        &                    0.18m, 8.88\degree &                   0.27m, 10.99\degree &                    0.42m, 15.12\degree &           0.37m, 20.48\degree &                    0.43m, 16.99\degree &          0.29m, 12.27\degree &                   0.35m, 11.32\degree &                   0.33m, 13.72\degree \\
      L3 (PL)                        &                    0.18m, 9.04\degree &                   0.27m, 10.84\degree &                    0.52m, 16.04\degree &           0.32m, 17.68\degree &                    0.45m, 18.82\degree &          0.29m, 11.63\degree &                   0.34m, 11.04\degree &                   0.34m, 13.58\degree \\
      L2 (PL)                        &                    0.15m, 7.80\degree &                    0.23m, 9.03\degree &                    0.30m, 14.41\degree &           0.20m, 16.44\degree &                    0.28m, 17.64\degree &           0.22m, 8.56\degree &                    0.28m, 9.37\degree &                   0.24m, 11.89\degree \\
      L1 (PL)                        &                    0.15m, 7.58\degree &                    0.22m, 8.74\degree &                    0.29m, 14.61\degree &           0.19m, 14.71\degree &                    0.28m, 14.98\degree &           0.21m, 7.78\degree &                    0.27m, 8.61\degree &                   0.23m, 11.00\degree \\
      L1 (PL+EA)                     &                    0.16m, 7.51\degree &                    0.25m, 8.40\degree &                    0.34m, 13.68\degree &           0.21m, 13.35\degree &                    0.27m, 14.64\degree &           0.21m, 8.16\degree &                    0.28m, 8.10\degree &                   0.24m, 10.55\degree \\
      L1 (PL+EA+H)                   &                    0.19m, 7.81\degree &                    0.27m, 8.94\degree &                    0.36m, 14.51\degree &           0.20m, 11.23\degree &                    0.31m, 17.69\degree &           0.23m, 8.41\degree &                    0.28m, 8.60\degree &                   0.26m, 11.03\degree \\
      L1 (PL+EF)                     &                    0.16m, 7.66\degree &                    0.23m, 8.99\degree &                    0.39m, 13.70\degree &           0.26m, 14.17\degree &                    0.28m, 13.73\degree &  \textbf{0.19}m, 7.08\degree &                    0.26m, 8.82\degree &                   0.25m, 10.59\degree \\
      L1 (PL+EF+H)                   &                    0.15m, 7.41\degree &                    0.23m, 9.06\degree &                    0.39m, 12.86\degree &           0.19m, 11.83\degree &           \textbf{0.26}m, 13.50\degree &           0.20m, 8.04\degree &                    0.25m, 9.00\degree &                   0.24m, 10.24\degree \\
      L1 (PL+PA)                     &                    0.17m, 7.14\degree &                    0.25m, 8.72\degree &                    0.34m, 15.38\degree &           0.20m, 12.07\degree &                    0.27m, 13.22\degree &           0.22m, 8.04\degree &                    0.27m, 9.10\degree &                   0.25m, 10.52\degree \\
      L1 (PL+PA+H)                   &                    0.15m, 6.86\degree &                    0.24m, 8.74\degree &                    0.31m, 12.19\degree &  0.17m, \textbf{11.15}\degree &                    0.27m, 13.68\degree &  \textbf{0.19}m, 7.53\degree &                    0.25m, 8.83\degree &                    0.23m, 9.85\degree \\
      L1 (PL+PF)                     &                    0.16m, 7.86\degree &                    0.25m, 8.65\degree &                    0.40m, 14.88\degree &           0.19m, 14.15\degree &                    0.28m, 12.91\degree &           0.20m, 7.65\degree &                    0.27m, 9.11\degree &                   0.25m, 10.74\degree \\
      L1 (PL+PF+H)                   &                    0.16m, 7.03\degree &                    0.23m, 8.36\degree &                    0.36m, 14.34\degree &  \textbf{0.14}m, 12.05\degree &                    0.29m, 12.81\degree &           0.21m, 8.58\degree &           \textbf{0.24}m, 8.62\degree &                   0.23m, 10.26\degree \\
      L1 (PL+ATL)                    &                    0.14m, 6.85\degree &                    0.23m, 7.71\degree &                    0.33m, 13.96\degree &           0.18m, 12.55\degree &                    0.27m, 12.88\degree &           0.20m, 7.63\degree &                    0.25m, 8.29\degree &                    0.23m, 9.98\degree \\
      L1 (PL+FTL)                    &                    0.15m, 7.12\degree &                    0.23m, 8.05\degree &                    0.28m, 15.49\degree &           0.20m, 12.89\degree &                    0.27m, 12.94\degree &           0.20m, 7.33\degree &                    0.25m, 8.25\degree &                   0.22m, 10.29\degree \\
      \bottomrule
    \end{tabular}
  }
\end{table*}